\documentclass[10pt,twocolumn,letterpaper]{article}

\usepackage{wacv}
\usepackage{times}
\usepackage{epsfig}
\usepackage{graphicx}
\usepackage{amsmath}
\usepackage{amssymb}

\usepackage{float}
\usepackage{stfloats}
\usepackage{subfigure}
\usepackage{multirow}
\usepackage{cuted}
\usepackage{capt-of}
\usepackage[ruled]{algorithm2e}
\usepackage{capt-of}

\def\wacvPaperID{453} 

\wacvfinalcopy 

\newcommand{\lego}{LEGO\textregistered\,}
\newcommand{\legos}{LEGOs\textregistered\,}

\ifwacvfinal
\usepackage[breaklinks=true,bookmarks=false,colorlinks]{hyperref}
\else
\usepackage[pagebackref=true,breaklinks=true,colorlinks,bookmarks=false]{hyperref}
\fi

\begin{document}

\title{Image2Lego: Customized \lego Set Generation from Images}

\author{Kyle Lennon, ~Katharina Fransen, ~Alexander O'Brien, ~Yumeng Cao, \\~Matthew Beveridge, ~Yamin Arefeen, ~Nikhil Singh, ~Iddo Drori\\
Massachusetts Institute of Technology\\ 
{\tt\small \{krlennon,kfransen,adobrien,ymcao,mattbev,yarefeen,nsingh1,idrori\}@mit.edu}
}

\maketitle

\begin{strip}
    \centering
    \includegraphics[width=\textwidth]{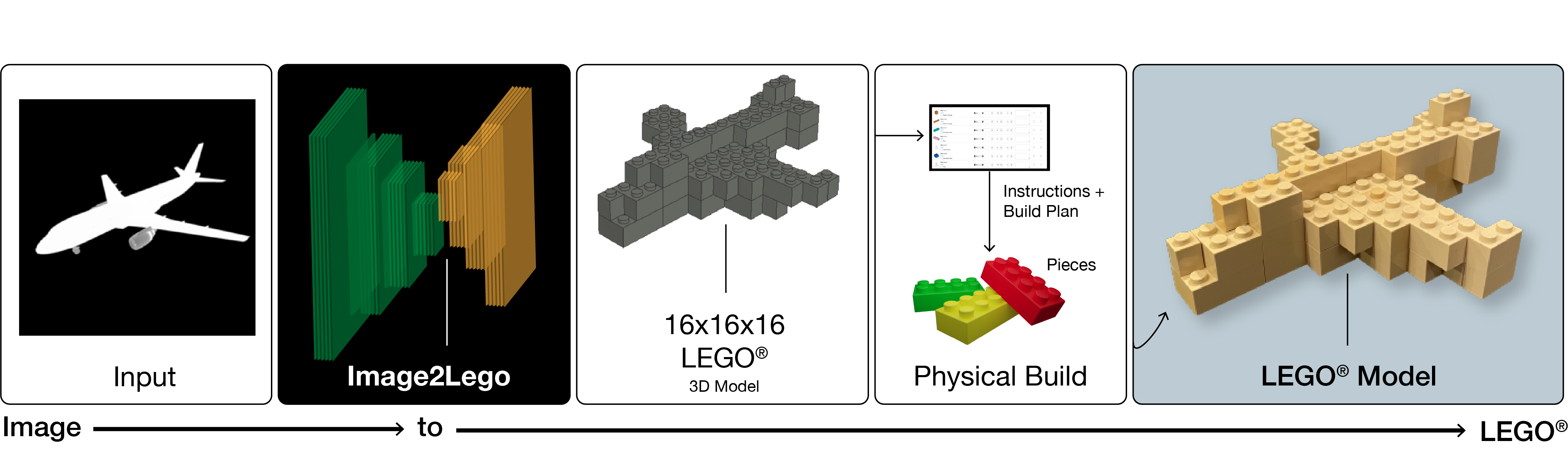}
    \captionof{figure}{Our method takes an image as input and produces a voxelized 3D model, which is then converted to a \lego brick set. From the provided pieces and instructions, the \lego model can then be built in the real world; example shown at the right.}
    \label{fig:airplane_intro}
\end{strip}

\begin{abstract}
\vspace{-9pt}
Although \lego sets have entertained generations of children and adults, the challenge of designing customized builds matching the complexity of real-world or imagined scenes remains too great for the average enthusiast. In order to make this feat possible, we implement a system that generates a \lego brick model from 2D images. We design a novel solution to this problem that uses an octree-structured autoencoder trained on 3D voxelized models to obtain a feasible latent representation for model reconstruction, and a separate network trained to predict this latent representation from 2D images. \lego models are obtained by algorithmic conversion of the 3D voxelized model to bricks. We demonstrate first-of-its-kind conversion of photographs to 3D \lego models. An octree architecture enables the flexibility to produce multiple resolutions to best fit a user's creative vision or design needs. In order to demonstrate the broad applicability of our system, we generate step-by-step building instructions and animations for \lego models of objects and human faces. Finally, we test these automatically generated \lego sets by constructing physical builds using real \lego bricks.
\end{abstract}

\section{Introduction}
For decades, \lego bricks have been a staple of entertainment for children and adults alike, offering the ability to construct anything one can imagine from simple building blocks. For all but the most exceptional \lego engineers, however, dreams quickly outgrow skills, and constructing the complex images around them becomes too great a challenge. \lego bricks are extraordinarily flexible by nature and have been assembled into intricate and fantastical structures in many cases, and simplifying the process of constructing the more complex designs is an important step to maintain appeal for amateur builders and attract a new generation of \lego enthusiasts. To make these creative possibilities accessible to all, we develop an end-to-end approach for producing \lego-type brick 3D models directly from 2D images. Our work has three sequential components: it (i) converts a 2D image to a latent representation, (ii) decodes the latent representation to a 3D voxel model, and (iii) applies an algorithm to transform the voxelized model to 3D \lego bricks. As such, our work represents the first complete approach that allows users to generate real \lego sets from 2D images in a single pipeline. A high-level demonstration of the full Image2\lego pipeline is presented in Figure \ref{fig:airplane_intro}, where a gray-scale 2D photograph of an airplane is converted to a 3D \lego model, and the corresponding instructions and brick parts list are used to construct a physical \lego airplane build. We tackle the issues specific to constructing high-resolution real 3D \lego models such as color and hallow structures. Our key contributions are:

\begin{itemize}
    \item A pipeline that combines creating a 3D model from a 2D image with an algorithm for mapping this 3D model to a set of \lego-compatible bricks, to provide this new Image2\lego application,
    \item An evaluation by examples and analysis to show how and when this pipeline works.
\end{itemize}

\begin{figure*}[!ht]
    \centering
    \includegraphics[width=\textwidth]{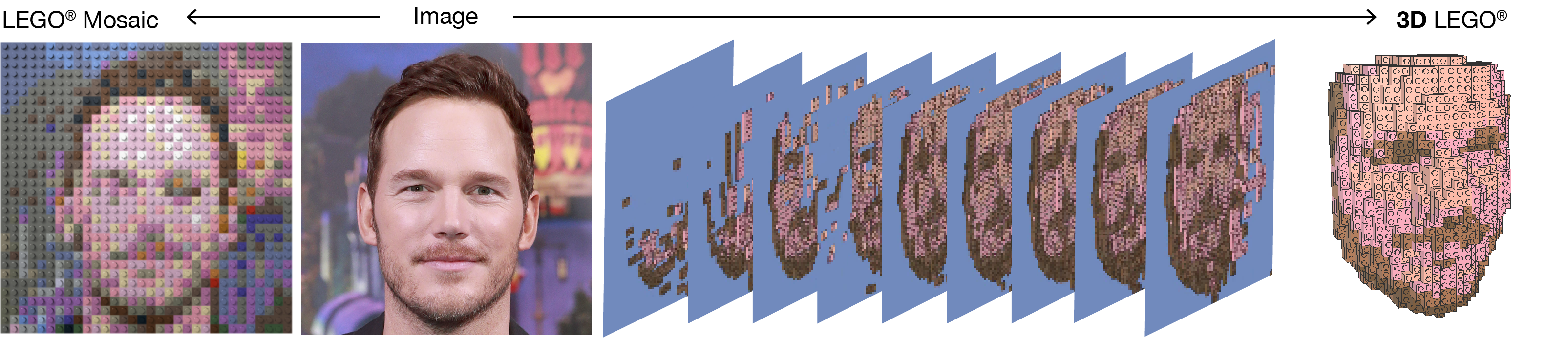}
    \caption{Center-Left: input image; Left: 2D Lego Mosaic output; Right: 3D \lego Model output and corresponding build process.}
    \label{fig:cp_build}
\end{figure*}

\begin{figure}
    \centering
    \includegraphics[width=\columnwidth]{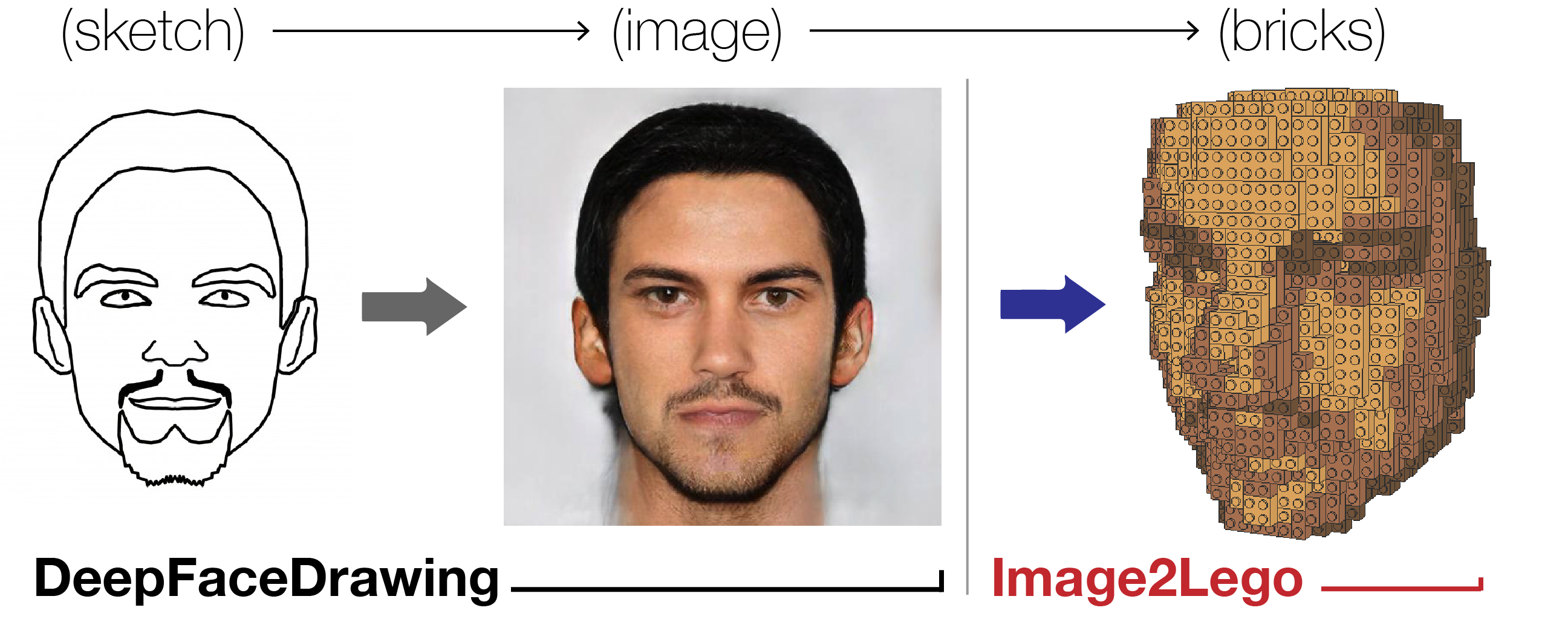}
    \caption{Left: input sketch; Center: 2D face image; Right: 3D face \lego.}
    \label{fig:sketch_example}
\end{figure}

Though we focus in this paper on our novel approach for multi-class object-image-to-lego construction, the same approach is extended to other creative applications by leveraging previous image-to-model work. For instance, generating \lego models from pictures of one's face is already an application of interest, but current work is limited to the generation of 2D \lego mosaics from images, such as that shown to the left of Figure \ref{fig:cp_build} (generated by the commercial product called \lego Mosaic Maker \cite{legomosaicmaker}). However, we extend the Image2\lego pipeline to include the pre-trained Volumetric Regression Network (VRN) for single-image 3D reconstruction of faces \cite{jackson2017vrn}. As shown in Figure \ref{fig:cp_build}, in contrast to the 2D mosaic, our approach generates a 3D \lego face from a single 2D image. Moreover, other learned tools may be appended or prepended to the base pipeline to develop more imaginative tools. For instance, by prepending the VRN with a sketch-to-face model \cite{chenDeepFaceDrawing2020}, we develop a tool that directly converts an imagined drawing into a \lego model, as demonstrated in Figure \ref{fig:sketch_example}, offering nearly limitless creative possibilities. In \S\ref{sec:caption2lego}, we demonstrate another extension, where we apply the Image2\lego pipeline with DALL-E~\cite{Ramesh2021} outputs to create a tool that automatically converts captions to \lego models.

\subsection{Related Work}

\paragraph{Image-to-3D}

3D model construction from 2D images of objects is an active research area \cite{Fu2021,Kniaz2020,Yu2021}. For example, Lim et al. \cite{Lim2013} demonstrate an algorithm for modeling fine-pose of objects within captured 2D images and matching them to a set of 3D models. GAN-based approaches \cite{pan20202d, hu2021self} for 3D reconstruction demonstrate high quality outputs and have recently been extended to allow control over the output. Girdhar et al. \cite{Girdhar2016} develop vector representations of 3D objects that are predictable from 2D images, and Stigeborn \cite{Stigeborn2018} develops machine learning methods for automatic generation of 3D models through octree-based pruning. The methods described in any of these previous works could feasibly be utilized for producing the desired 3D models for the goals of this work, though the end goal of a \lego model has not been previously considered and requires adaptations, which we explore in this work (e.g. image to flexible-resolution voxelized 3D models), to create models optimal for creative \lego engineering.

\paragraph{3D-to-\lego}
The challenge of converting voxelized 3D models into \lego designs has been previously explored as well. Silva et al. \cite{Silva2009} demonstrate real-time conversion of surface meshes to voxels to \legos, and Lambrecht \cite{Lambrecht2006} describes methods for high-detail \lego representations of triangle mesh boundaries. However, a gap has remained between 3D model generation from images and \lego generation from 3D models. The goal of our work is to bridge this gap by developing a complete Image2\lego pipeline, allowing anyone to create custom \lego models from 2D images.

The problem of \lego generation from images adds an additional goal to just 3D model generation, namely that it is important to have flexibility in the output resolution. Additionally, the latent space should have some flexibility so as to generate unseen structures from new input images. The former is useful in providing users with \lego designs of different scales and resolutions, to better achieve varying levels of difficulty, availability of material resources, and cognitive effort. For instance, small renditions of an object may be useful as fine elements in a greater scene, while larger renditions may serve as independent \lego models. The latter feature of a generalizable latent space allows users to generate new \lego sets that are associated with new captured images. This work represents the first effort to combine these approaches, using an octree-structured autoencoder in the image-to-model pipeline. We evaluate its ability to perform this task on new images in several examples.

\section{Methods}

Since the VRN, which we use for faces, is pre-trained, the majority of this section is focused on the construction and training of the TL-Octree network (applied to object images). This network is inspired by the Octree Generating Network \cite{Tatarchenko2017}, which consists of a convolutional block in an `octree' structure that sequentially up- (or down-) samples the resolution of a 3D model by a factor of two in each spatial dimension, and the TL-Embedding network \cite{Girdhar2016}, which accomplishes image-to-3D reconstruction by simultaneously training an autoencoder to compress the 3D model representation into a latent vector and an image encoder to predict the latent representation of a model from a single 2D image. Our model combines these two approaches, as shown in Figure \ref{fig:architecture}. Specifically, an autoencoder is trained on multiple classes of 3D models, and the latent representations of this autoencoder are taken as the targets for a separate 2D image encoder. At testing time, the `encoder' portion of the autoencoder is discarded, such that an image is fed through the image encoder and the `decoder' of the autoencoder to reconstruct the 3D model.

\subsection{Dataset}

The TL-Octree network is trained and tested on the ModelNet40 dataset~\cite{Wu2015}, a collection of 40 categories of 3D meshes composed of 8,672 models for training and 2,217 models for testing. The surfaces of these meshes are sampled to generate point clouds, and these point clouds are subsequently converted to voxels at a resolution of $32^3$. To obtain paired 2D images, we render the 3D models in a set pose using Pyrender \cite{pyrender}. We focus on model reconstruction from a single, set pose here to demonstrate a simple working example of our Image2\lego pipeline, but note that the pipeline may be readily extended to 3D reconstruction from different or multiple poses by incorporating ideas from previous work in this area \cite{Choy2016}.

\subsection{3D-Model Autoencoder}
\begin{figure*}[!ht]
    \centering
    \includegraphics[width=\textwidth]{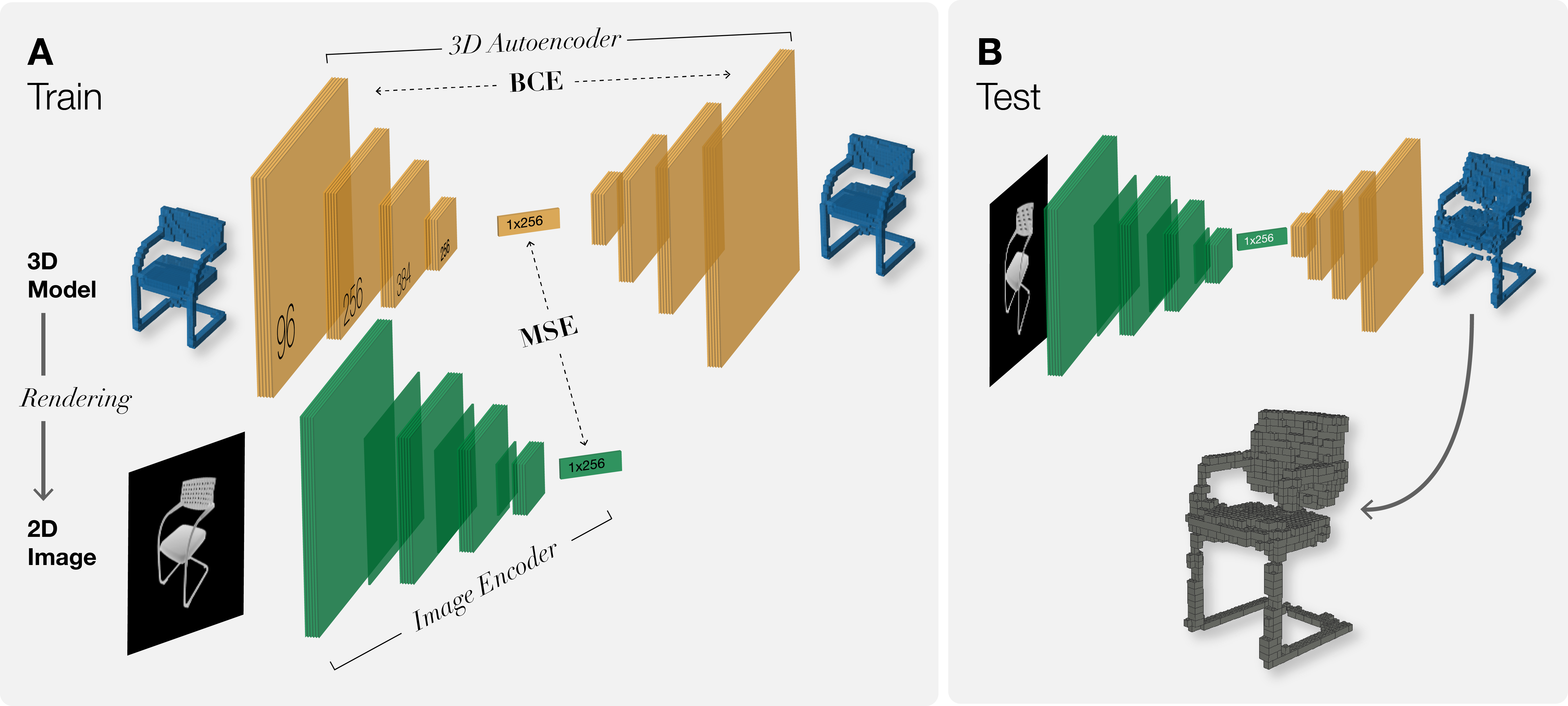}
    \caption{Our TL-Octree network, consisting of a 3D model autoencoder and an image encoder. At train time, we optimize the autoencoder weights for reconstruction of the voxelized 3D models. We additionally minimize the mean squared error between the latent representations of this encoder and those produced by the image encoder from a corresponding 2D image. At test time, we compose the image encoder with the 3D decoder network to produce a voxelized 3D model from an image. Finally, we algorithmically produce a matching \lego approximation.}
    \label{fig:architecture}
\end{figure*}

We begin by training an autoencoder on the voxelized 3D models. The decoder of this network is adapted from the Octree Generating Network \cite{Tatarchenko2017}. It consists of multiple blocks, each of which upsamples the resolution by a factor of two in each dimension (i.e. an octree structure) and changes the number of channels, beginning with the latent representation, parameterized as a 256-dimensional vector (i.e. a single voxel with 256 channels). Below, we discuss a brief hyperparameter optimization campaign to determine the number of channels in each subsequent layer after the latent representation. The resolution upsampling is accomplished by generative convolution transpose layers, which efficiently parse and prune the sparse structure of the input tensor to minimize memory usage and runtime \cite{gwak2020generative}. Our encoder mirrors the decoder, replacing each generative convolution transpose with a conventional stride-2 convolutional layer. In both the encoder and decoder, each convolutional layer (or generative convolutional layer) is followed by a batch normalization layer to reduce overfitting, and an Exponential Linear Unit (ELU) activation. This autoencoder architecture is shown in Figure \ref{fig:architecture}A.

We implement the layers and sparse tensor storage for voxelized objects using the Minkowski Engine \cite{Choy2019}, which implements sparse tensor objects and sparse tensor layers. Critically, employing this sparse tensor framework makes our network generalizable to higher 3D model resolutions without encountering severe runtime or memory restrictions, whereas other models that represent the 3D objects densely are limited to low resolutions. For a single example object, the loss is computed by summing the cross-entropy between the sigmoid activation of the decoded object across every decoding step (i.e. at resolutions $4^3$, $8^3$, $16^3$, and $32^3$), with the corresponding resolution discretization of the input object:
\begin{equation}
    \mathcal{L}_{AE} = -\sum_{r=1}^{N}\sum_{n=1}^{N_r}\left[y_n\ln\sigma(x_n) + (1 - y_n)\ln(1 - \sigma(x_n))\right]
\end{equation}
where $N$ is the total number of resolutions output by the model (in this case, $N = 4$), $N_r$ is the number of voxels in the output of an object at resolution $r$, $x_n$ is the pre-activated output at a specific voxel location, $\sigma$ is the sigmoid activation, and $y_n$ is the corresponding target voxel status (1 for filled, 0 for unfilled). Thus, from the latent representation we may recover voxelized objects of four different resolutions while only training the network a single time, meeting the design criterion of resolution-flexibility set forth previously. We train this network using stochastic gradient descent with a mini-batch size of 16.

\begin{figure*}[!t]
    \centering
    \includegraphics[width=\textwidth]{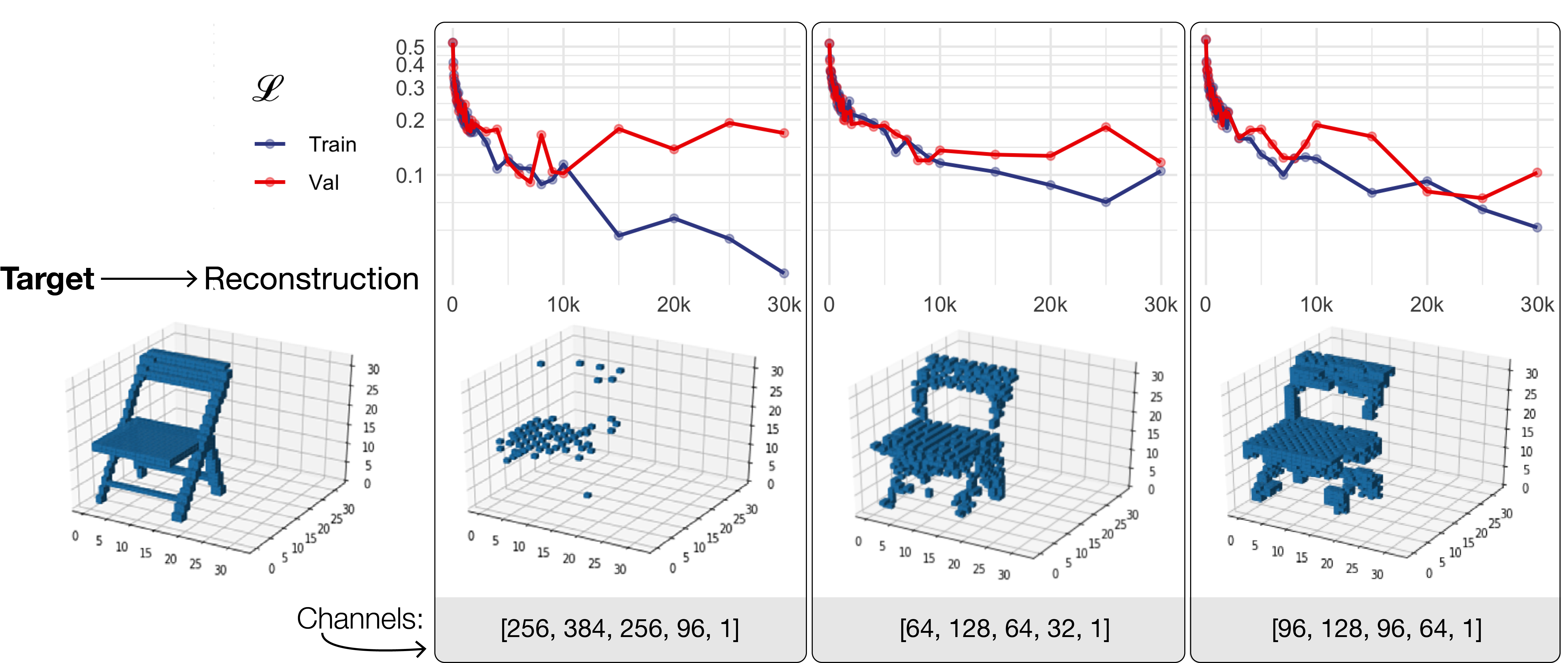}
    \caption{Depiction of Autoencoder training and validation loss as a function of iteration number for different channel numbers. An input image and reconstructed output for each architecture is depicted for visual interpretation of the loss.}
    \label{fig:optimization}
\end{figure*}

We evaluated our training procedure with different sized networks to examine how this affected downstream performance. These results are plotted in Figure \ref{fig:optimization}, for decoders with the following number of channels on each layer: $[265, 384, 256, 96, 1]$, $[64, 128, 64, 32, 1]$, and $[96, 128, 96, 64, 1]$, chosen to be similar to the network size of the original TL-Embedding network \cite{Girdhar2016}. We observe the largest network overfitting to the training data, resulting in worse test performance. While performance is similar between the two reduced-size networks, the folding chair reproduction in the medium-sized network is qualitatively slightly improved (the smallest network appears to introduce rough-edge artifacts). As such, we use the $[96, 128, 86, 64, 1]$ network.

\subsection{2D Image Encoder}
Next, we train a separate encoder network to predict the 256-dimension latent representations learned from our 3D autoencoder using the rendered single 2D images of the objects. Our encoder design is based on AlexNet \cite{Krizhevsky2012}, a seminal network in image classification. Starting from single-channel gray-scale images with resolution $128^2$, convolutional layers increase the number of channels to 96, 256, 384, and 256, while intermittent max pooling layers decrease the resolution by a factor of two each time. Finally, fully-connected layers convert the 256 channel $4^2$ tensor to a 256-dimension feature vector. For optimization, we compute the mean-squared loss with the corresponding autoencoder latents: $\mathcal{L}_{enc} = \frac{1}{n}\sum_{n}(y_n - x_n)^2$, where $x_n$ represents the predicted latent feature on dimension $n$, and $y_n$ represents the target latent feature.

This network is trained using the Adam optimizer with a mini-batch size of 128 for consistency with prior work. The fully trained image encoder and 3D autoencoder implement the first two steps of our Image2\lego pipeline. As depicted in Figure \ref{fig:architecture}, the image encoder takes an input image and produces a feature vector, which is fed to the 3D decoder network. As such, at test time, we do not need the 3D encoder network.

\subsection{Voxel to \lego Conversion}

The final step is to convert voxelized 3D models to \lego builds. This is done deterministically with an algorithm adapted from the ColouredVoxels2Lego project \cite{Marsden2019}. This algorithm first converts each voxel to a 1 by 1 \lego brick, and then iterates through each layer in the $z$-coordinate of the model to find optimal groups of bricks to be combined into larger cuboid bricks. The output of this software is a written LDraw file, which may be visualized in a software such as LeoCAD~\footnote{\href{https://www.leocad.org/}{https://www.leocad.org/}}. For colored voxels, the original implementation of this algorithm matched each voxel to a \lego brick of existing color by finding the \lego color that minimizes the Euclidean distance to the voxel color in RGB space. However, for models that include color, we have adapted the handling of color to limit excessive variations in the face of a continuous color gamut in the 3D models, which will be discussed in the following section. By appending this adapted algorithm to the output of the 3D model decoder, we obtain our full Image2\lego pipeline.

\subsection{Color Quantization}

In certain application of the Image2\lego pipeline, such as for \lego models of faces, color is a critical feature. Although color is not handled in the TL-Octree network, which focuses on shape reconstruction, it is required for the \lego reconstruction of faces using the VRN. The 3D objects generated by the VRN contain a continuous gamut of colors, which, when fed to the original implementation of the voxels-to-\lego algorithm, may produce a patchiness due to the discrete and rather limited gamut of \lego brick colors. To increase the uniformity of color in the \lego products, we prepend the voxels-to-\lego algorithm with a $k$-means clustering algorithm, which determines from the span of colors in the 3D object the $k$ predominant colors and subsequently assigns each voxel to the nearest predominant color in RGB space (with a Euclidean distance metric). These $k$ predominant colors are then mapped to the nearest \lego brick color. For the examples shown in this work, $k = 4$, which was qualitatively found to sufficiently distinguish important features of the face (e.g. eyes, nose, mouth, facial hair) without producing excessive variations in color across the skin.

\subsection{Hollow Shell vs. Full 3D Model}
\begin{figure}
    \centering
    \includegraphics[width=\columnwidth]{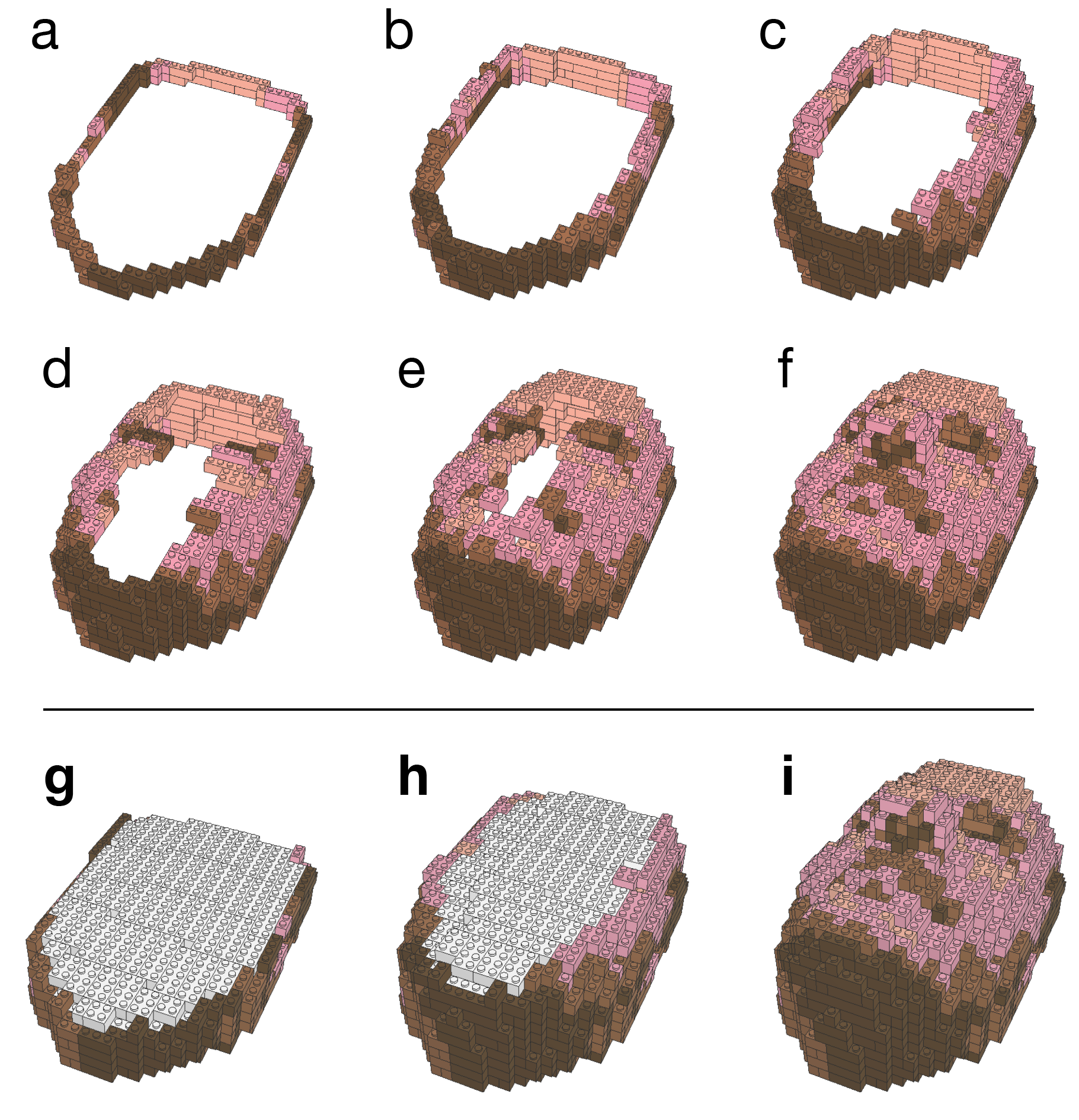}
    \caption{A sample of \lego build instruction steps with an image of a face. Top: The internal structure is hollow, which leads to practical build challenges. Bottom: to mitigate this, we modify the original algorithm to fill in the structure with additional bricks.}
    \label{fig:3dface-shell}
\end{figure}

For 3D faces, our original models were hollow, resembling a shell as shown in Figure \ref{fig:3dface-shell} for \lego of a 3D face. We found that a hollow structure is difficult to physically build since the model often collapses inwards when built on a \lego surface. We therefore modified our original algorithm to fill in the 3D structure with bricks, rather than generating a hollow shell.

\section{Results}

We make our models and code for training and running our pipeline available on the project web page.\footnote{Project~page:~\href{https://krlennon.github.io/image2lego/}{https://krlennon.github.io/image2lego/}}

\subsection{Learned Model Performance}

The autoencoder was evaluated quantitatively using the Jaccard index, or intersection-over-union. Intersection-over-union (IoU) loss is a well established loss for 3D object detection\cite{10.1007/978-3-030-58565-5_28}, image segmentation\cite{Rezatofighi_2019_CVPR}, and pixelwise prediction \cite{pmlr-v139-yu21e} with the advantage of scale invariance. We calculate the test performance of the autoencoder using the IoU score between the model and reconstruction $\mathrm{IoU} = \frac{|X \cap Y|}{|X \cup Y|}$, where $X$ represents the set of filled voxels in the prediction, $Y$ the set of filled voxels in the target, and $|.|$ the cardinality of a set. These scores are listed in Table \ref{tab:iou} for the chair and airplane categories of the ModelNet40resolutions. The results show high fidelity reconstructions, with accuracy greater than 70\% up to $16^3$ resolution. This performance is comparable to that in previous studies \cite{Tatarchenko2017}, with the slightly lower reconstruction accuracy likely attributable to the fact that our model has been trained on many object classes, whereas many previous studies focus on single classes. The accuracy falls at $32^3$, which may be attributed to the decreased ability of the autoencoder to capture details on the object surface, compared to general features of the object shape and size. We note the Image-to-Model IoU scores are consistently within $\sim 0.2-0.3$ of the reconstructions, with performance decreases due primarily to the more limited representational capacity of a 2D image compared to a 3D model.

\begin{table*}[!t]
    \centering
    \begin{tabular}{c|cccc|cccc}
         & \multicolumn{4}{c}{Autoencoder IoU} & \multicolumn{4}{c}{Image-to-Model IoU} \\
        Resolution & $4^3$ & $8^3$ & $16^3$ & $32^3$ & $4^3$ & $8^3$ & $16^3$ & $32^3$ \\\hline
        Chair & 0.972 & 0.870 & 0.700 & 0.465 & 0.796 & 0.633 & 0.455 & 0.240 \\
        Airplane & 0.985 & 0.868 & 0.705 & 0.439 & 0.749 & 0.528 & 0.431 & 0.228 \\\hline
    \end{tabular}
    \vspace{0.2cm}
    \caption{Intersection-over-union scores for the voxelized 3D model reconstructions at each resolution by the 3D autoencoder and the full image-to-model pipeline (the image encoder prepended to the decoder of the 3D autoencoder).}
    \label{tab:iou}
\end{table*}

Two particular challenges evident in the 3D reconstruction outputs are the models limited ability to construct fine details regarding depth, and it's limited ability to anticipate the shape of occluded parts of the model in the rendered images. These challenges are typical of 3D model reconstruction from single images \cite{Choy2016}. The former is typically reflected in rough surface features, and the latter reflected in missing voxels, for example in the partially or totally occluded back legs of chairs. However, we note that these challenges may be remedied by extending the Image2\lego pipeline to include 3D model reconstruction from multiple images of a single object, which we leave to future work.

\subsection{Image2\lego Examples}
\begin{figure*}[!t]
    \centering
    \includegraphics[width=0.9\textwidth]{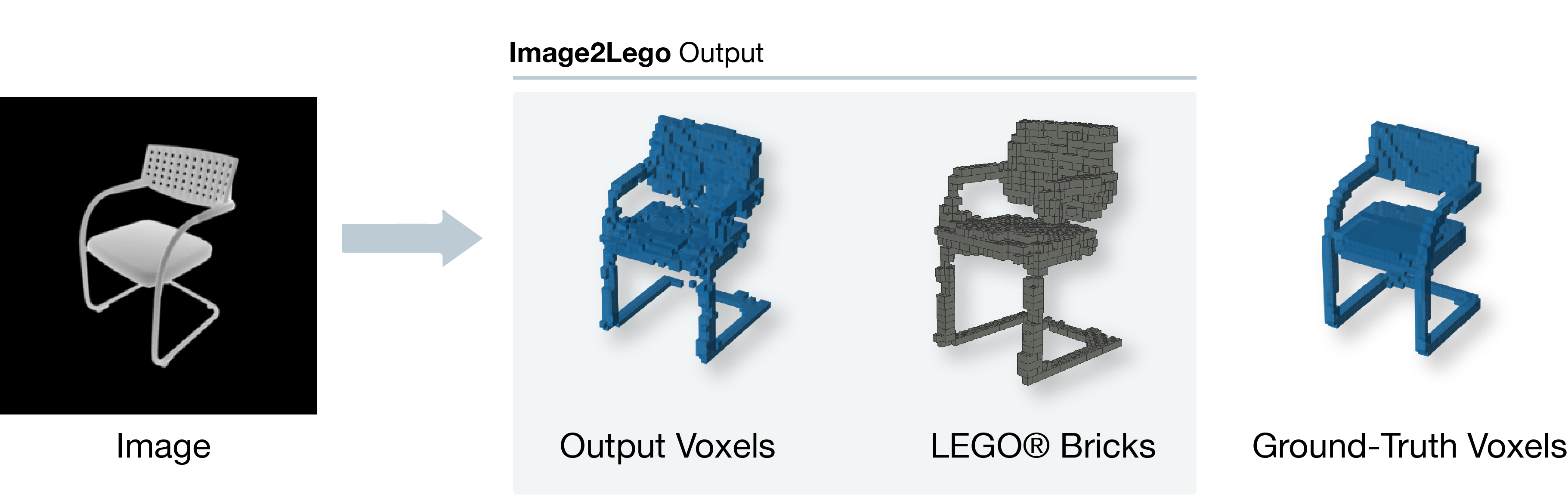}
    \caption{Example results: an input image that our model renders as voxels, from which we then produce a \lego version (target voxelized object shown at right).}
    \label{fig:chair}
\end{figure*}

\begin{figure*}[!t]
    \centering
    \includegraphics[width=\textwidth]{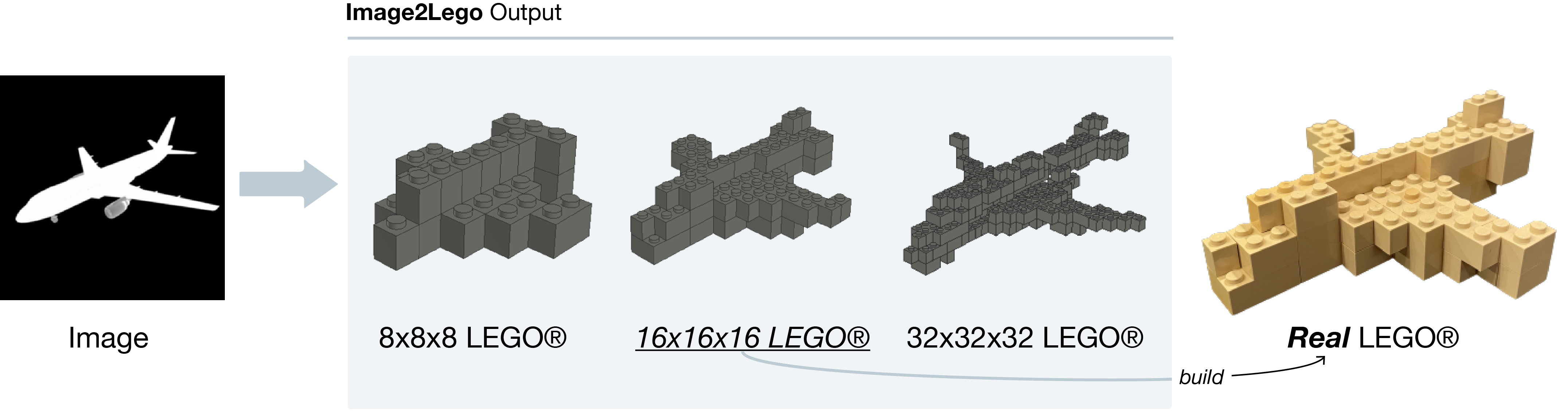}
    \caption{Example results: variable resolution \lego arrangements from one given gray-scale image of an airplane. The mid-resolution airplane is physically build using real \lego bricks, as shown on the right.}
    \label{fig:plane}
\end{figure*}

Figure \ref{fig:chair} demonstrates the steps of the completed pipeline for a chair from the test data set. The output model at maximum resolution is qualitatively similar to the target model -- specifically, it shares the same structure of legs, arms, and chair back, as well as similar size and aspect ratio. The reconstruction is clearly imperfect, as demonstrated by the surface roughness. Upon close examination, one notices that the final \lego model consists of bricks of various shapes, as a result of the conversion of voxels to \lego bricks. The inclusion of many brick shapes leads both to a more interesting building experience, and to enhanced structural stability of the \lego model. However, the \lego structure has not been quantitatively evaluated for stability, and therefore may not in all cases be physically realizable without further modifications. Future work might consider modifying current algorithms with stability-inducing heuristics, as previously explored in other works studying 3D model to \lego conversion \cite{luo2015legolization}.

To demonstrate another important feature of our pipeline, we use a single image to obtain \lego models of various sizes. Figure \ref{fig:plane} demonstrates the multiple-scale resolution capabilities of our model for an example airplane image, also from the test data set. \lego models are generated at $8^3$, $16^3$, and $32^3$ resolution (the $4^3$ model for this example is too low-resolution to be identified as a plane, so it is omitted here). At all resolutions higher than $4^3$, the plane-like structure is evident. These resolutions are able to meet varying user needs. For instance, a smaller model may be useful as a part of a larger \lego scene, while the larger models may be useful for stand-alone sets. At the lower two resolutions, the model is fully connected and represents a physically realizable build. To this end, we have assembled a list of brick pieces needed to construct the $16^3$ \lego airplane (a total of 40 bricks), and automatically generated a set of building instructions in LeoCAD.

\begin{figure*}[!t]
    \centering
    \includegraphics[width=\textwidth]{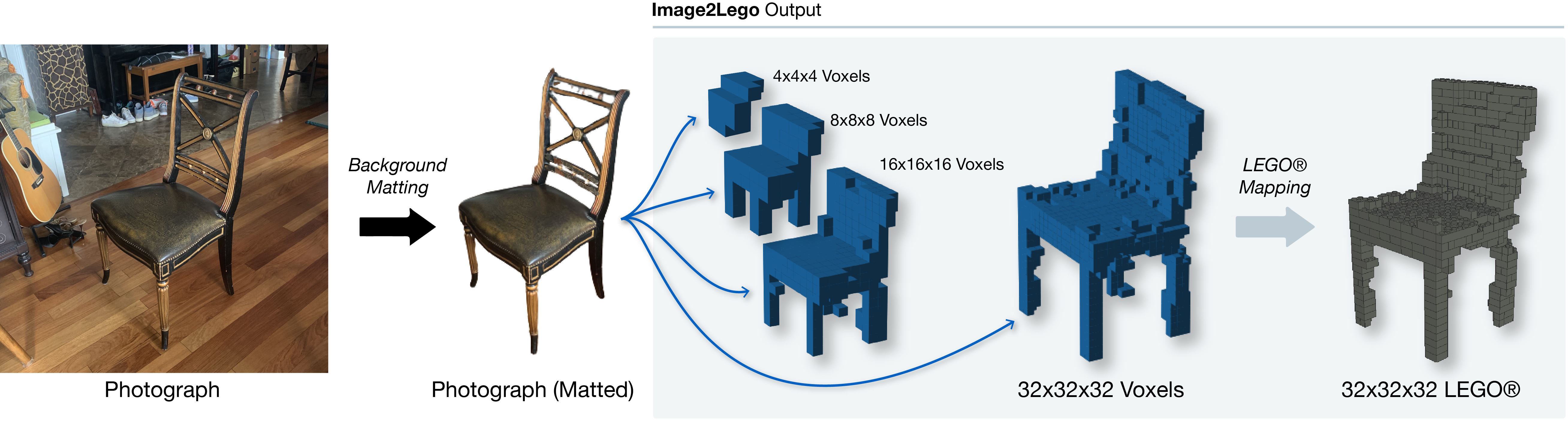}
    \caption{The full pipeline of our model: we input a photograph and apply background matting first, as a preprocessing step. Our model then takes this as input and produces a 3D voxel grid. We then supply this to an algorithm that maps it to a \lego structure. The output resolution is flexible, offering opportunities to construct plausible \lego models at different levels of effort, expertise, and availability of bricks for construction.}
    \label{fig:chair-intro}
\end{figure*}

We next demonstrate the capabilities of our model in the real world, with user-supplied images. Figure \ref{fig:chair-intro} presents the full pipeline for a real photograph captured with one author's smartphone, which, to our knowledge, represents the first ever example of a \lego model produced directly from a photograph, through a combination of background matting \cite{sengupta2020background,lin2021real} and our Image2\lego method. This is a significant step forward for hopeful hobbyists, and we expect that it could lead to an easily implementable and highly accurate program once trained with a larger and more varied data set in the future.

\begin{figure*}[!t]
    \centering
    \includegraphics[width=0.9\textwidth]{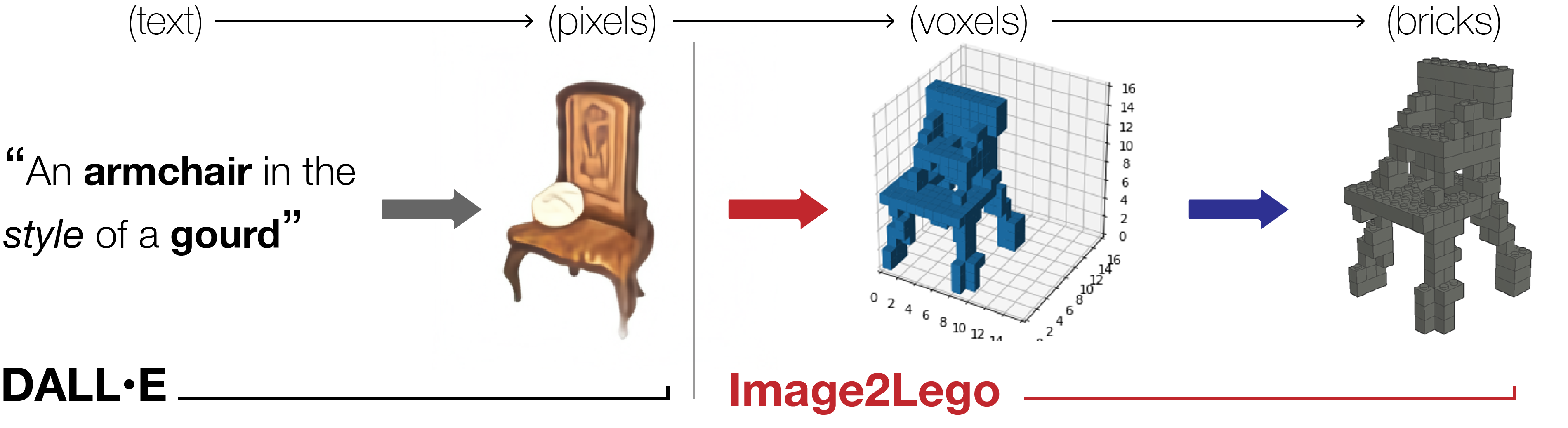}
    \caption{An example representation of the capabilities for combining our system with DALL-E to produce \lego models from written descriptions. The DALL-E model produces an image of a chair based on a description provided by the user, following which the image of the chair is converted to a 3D \lego model by our program. }
    \label{fig:dalle}
\end{figure*}

\subsection{Caption2\lego}
\label{sec:caption2lego}

As discussed in the Introduction, the Image2\lego pipeline is amenable to various extensions, be they different adaptations of the image-to-3D model architecture or pre- or post-procesing techniques that may enhance the input or output space of the pipeline. One example that we consider here is to add an initial caption-to-image conversion step, accomplished by the DALL-E model~\cite{Ramesh2021}, which produces images from arbitrary written descriptions. This extension of the pipeline dramatically expands the creative space for users by not limiting \lego designs to those that can be easily photographed, but rather to any fictional or real object that can be described in words. We present an example in Fig.~\ref{fig:dalle}, with the input caption: "An armchair in the style of a gourd" -- an object that may or may not exist to be photographed in the real world, but which DALL-E can produce as a photorealistic rendering. For all examples, see our project page.

\section{Conclusion}
We present a pipeline for producing 3D \lego models from 2D images. This work represents a significant advancement in the quest to generate physical \lego models from single 2D images. Our newly-trained TL-Octree structured network is able to construct the defining geometric features of multiple classes of objects from two-dimensional input images. We also demonstrate the pipeline extension to other, pre-trained or differently structured networks, such as the VRN for 3D face reconstruction, which generates colorful \lego models of a variety of object classes. In the future, these capabilities may be further improved with an expanded data set, which may include multiple object poses to mitigate the current limitations related to occluded spaces and fine depth features, and by incorporating additional features in the pipeline to make builds more usable and sophisticated (e.g. ensuring stable structures and expanding the catalog of \lego piece shapes beyond cuboid bricks). We believe this work will greatly improve the accessibility of creative, personalized LEGO creations for builders of all skills and all ages.

{\small
\bibliographystyle{ieee_fullname}
\bibliography{biblio}
}

\end{document}